\documentclass[runningheads]{llncs}
\usepackage[T1]{fontenc}
\usepackage{graphicx}
\usepackage{tabulary}
\usepackage{float}
\usepackage{amsmath, amssymb, mathtools}
\usepackage{subcaption}

\begin{document}
\title{Free Energy in a Circumplex Model of Emotion}
%
%
\author{Candice Pattisapu\inst{1}$^*$
    \and Tim Verbelen\inst{1}$^*$
    \and Riddhi J. Pitliya\inst{1,2}
    \and Alex B. Kiefer\inst{1,3}
    \and Mahault Albarracin\inst{1,4}}
\authorrunning{C. Pattisapu et al.}
\institute{VERSES Research Lab, Los Angeles, California, 90016, USA \and
Department of Experimental Psychology, University of Oxford, Oxford, UK \and
Department of Philosophy, Monash University, Melbourne, VIC, Australia \and
Department of Computer Science, Universit\'{e} du Qu\'{e}bec \`{a} Montr\'{e}al, Montr\'{e}al, Canada \\
\email{candice.pattisapu@gmail.com}}

%
%
%
\maketitle              

\def\thefootnote{*}\footnotetext{These authors contributed equally to this work}\def\thefootnote{\arabic{footnote}}

\begin{abstract}
 Previous active inference accounts of emotion translate fluctuations in free energy to a sense of emotion, mainly focusing on valence. However, in affective science, emotions are often represented as multi-dimensional. In this paper, we propose to adopt a Circumplex Model of emotion by mapping emotions into a two-dimensional spectrum of valence and arousal. We show how one can derive a valence and arousal signal from an agent's expected free energy, relating arousal to the entropy of posterior beliefs and valence to utility less expected utility. Under this formulation, we simulate artificial agents engaged in a search task. We show that the manipulation of priors and object presence results in commonsense variability in emotional states. 

\keywords{active inference \and emotional inference  \and circumplex model of emotion}
\end{abstract}
\section{Introduction}

Emotions are internal states that influence behavior and cognition~\cite{MartnezMiranda2005}. A comprehensive account of intelligence requires a theory of emotion and its relationship to higher-level cognitive processes, which is the main aim of this paper. We begin by briefly reviewing foundational accounts in psychological and active inference literature.

\subsection{Psychological Accounts of Human Emotion}

Two approaches to the taxonomy of emotions polarize scholarship on emotions. One class of theories, of which Ekman’s basic emotions model is the most mainstream~\cite{Ekman2011}, views emotions as discrete states. Here, it is proposed that a small set of emotion categories are the building blocks for more complex emotional states. On this account, anger, joy, disgust, fear, sadness, and surprise are core categorical elements of human emotional life. Imaging studies cited in support of this view associate discrete emotions with various brain structures. For example, fear has been correlated with activity in the amygdala~\cite{Lindquist2012}. 

In contrast, dimensional approaches aim to characterize emotions in terms of their relative locations in a continuous emotional state space (though these emotions may be discretized for practical applications). The most compelling dimensional approaches are premised on the Circumplex Model~\cite{Russell1980}, which characterizes emotions as mental states organized along orthogonal dimensions of valence and arousal. 

From the dimensional view, there are fluid boundaries between emotions. For example, when valence remains low and arousal diminishes, anger transitions to displeasure. Imaging studies associate arousal with the amygdala and valence with the orbitofrontal cortex~\cite{Lindquist2012}. This localization of valence and arousal dovetails with their functional roles: Arousal amounts to sensitivity to sensations, whereas valence involves appraisal of those sensations relative to the goals of an agent. Thus, the arousal-valence distinction maps onto the independently motivated distinction between low-level sensation and cognitive monitoring.  

Approaches to characterizing emotions as discrete cite the evolutionary plausibility of their theories~\cite{Plutchik1980}. Additionally, there is cross-cultural evidence supporting the existence of basic emotions~\cite{Russell1991}. However, there is a lack of consensus about which basic emotion categories exist, a debate that may be attributed to a lack of agreement on criteria about what a discrete emotion is in the first place~\cite{Ortony2021}. Indeed, discrete models have a difficult time explaining ``edge cases'' that are naturally accommodated on a dimensional view, such as the finding that fear-based arousal may be interpreted as attraction~\cite{DuttonAron1974}. Discrete emotions can be represented within a Circumplex model as attractor states in a continuous landscape. In addition, dimensional models are preferrable because they focus explicitly on the granularity of emotional state space, and, moreover, are the explanandum of most active inference formulations of emotion. For these reasons, we embrace dimensional models in what follows.  

\subsection{Previous Active Inference Formulations of Emotions}

In this section, we summarize existing research that has treated certain aspects of emotion within the active inference framework. While these models address elements such as valence, they do not provide comprehensive accounts. Notably, the dimension of arousal has been largely overlooked. Our formulation aims to address this gap by integrating both valence and arousal into a more comprehensive model of emotions under active inference.

Joffily and Coricelli~\cite{Joffily2013} cast valence as the interaction between the first- and second-order time derivatives of free energy, serving as an indicator of emotional well-being. In their framework, emotional valence provides feedback on an agent's learning process. A rapid decrease in free energy, indicating increasingly accurate predictions, is associated with happiness and suggests that the agent should update its models more quickly by increasing the learning rate. Conversely, negative valence suggests that a slower learning rate is appropriate. This adaptive mechanism regulates the rate of evidence accumulation to optimize learning. In one-armed bandit task simulations, agents that used emotional valence to adjust their learning rates better estimated statistical regularities in more volatile environments. There was a rapid increase and subsequent decline in negative emotional states in the presence of volatility, and agent performance was enhanced by leveraging this positive and negative valence.

Hesp et al.~\cite{Hesp2021} proposed a hierarchical active inference model of emotional valence. In their approach, valence is defined as changes in the expected precision of the action model, which they term ``affective charge''. Here, precision refers to confidence in the agent’s predictions and actions, which  may be regarded as an internal estimate of model fitness. Lower-level state factors, such as sensory inputs, are used to inform higher-level valence representations (``beliefs about beliefs''), which in turn influence the precision of potential action policies. Simulation studies using a T-maze paradigm showed that positive valence led to riskier behavior, interpreted as increased confidence in action. When the reward location was changed, resulting in negative valence, the agent displayed more conservative exploration, indicating a shift in the confidence and model adjustment. While the models proposed by Joffily and Coricelli~\cite{Joffily2013} and Hesp et al.~\cite{Hesp2021} focus primarily on valence, we are inspired by their accounts and propose a more comprehensive approach that also incorporates arousal to fully capture the emergence and dynamics of emotional inference. 

It is worth mentioning that Smith et al.~\cite{Smith2019} also explored emotional state inference under active inference. In this work, a combination of exteroceptive (external sensory), proprioceptive (body position), and interoceptive (internal sensory) observations are used to infer emotional states, given a hierarchical model in which valence and arousal are presupposed as part of the lower-level observation space. The focus in this work is on the learning of explicit, consciously accessible discrete representations of emotional states, based on explicit feedback supplied by a teacher/experimenter, rather than on the factors that constitute emotional states themselves and the ways in which these states modulate behavior. Therefore, this work does not directly inform our approach.

The aim of this paper is rather to provide a comprehensive account of emotional states themselves, formalizing the Circumplex Model of emotion by mapping both arousal and valence to aspects of free energy minimization within the active inference framework. We proceed as follows: In Section 2, we motivate mappings of valence and arousal to specific terms within the free energy functionals of active inference, and we demonstrate a transformation of the resulting valence and arousal dimensions of emotion into the space proposed by the Circumplex Model. In Section 3, we describe the setup of our simulation study, in which an artificial agent is tasked with finding an object in various scenarios. In Section 4, we present the results of our simulations, and in Section 5, we discuss their implications for understanding and formalizing emotional inference and behavior.

\section{From Free Energy to a Circumplex Model of Emotion}

The Circumplex Model of emotion organises emotional states along two dimensions: valence and arousal. We hypothesize that both dimensions can be derived from an active inference agent's free energy levels. 

\subsection{Active Inference}

Active inference casts perception and action as Bayesian inference~\cite{ActInfBook}, where an agent entertains a generative model of its environment and perceives and acts in order to minimize free energy, as defined below, with respect to this model. In general, such an agent's generative model can be written as the joint probability distribution over states $s$ and observations $o$. Minimizing free energy, then, entails finding the approximate posterior distribution $Q(s|o)$:

\begin{equation}
\begin{aligned}
\min_{Q(s|o)} F & = \underbrace{D_{KL}[Q(s|o)||P(s|o)]}_\text{posterior approximation} -\underbrace{\log P(o)}_\text{log evidence}\\
 & = \underbrace{-\mathbb{E}_{Q(s|o)}[\log P(o, s)]}_\text{energy} - \underbrace{H[Q(s|o)]}_\text{entropy} \\
 & = \underbrace{D_{KL}[Q(s|o))||P(s)]}_\text{complexity} - \underbrace{\mathbb{E}_{Q(s|o)}[\log P(o| s)]}_\text{accuracy}
\end{aligned}
\label{eq:F}
\end{equation}

Effectively, by minimizing free energy, an agent aims to find the model that maximizes accuracy with minimal complexity, which is effectively maximizing (a lower bound) on the model (log) evidence.

To interact with the environment, an agent in addition also needs to select a sequence of actions or policy $\pi$ to execute. In active inference, agents select policies that minimize expected free energy $G$:

\begin{equation}
\begin{aligned}
P(\pi) & = \sigma(- G(\pi)) \text{, with} \\
G(\pi) &= \sum_{\tau=t+1}^{T} \underbrace{\mathbb{E}_{Q(o_\tau|\pi)}\big[D_{KL}[Q(s_\tau|o_\tau,\pi)||Q(s_\tau|\pi)]\big]}_{\text{(negative) information gain}} - \underbrace{\mathbb{E}_{Q(o_\tau|\pi)}\big[\log P(o_\tau | C) \big]}_{\text{expected utility}}
\end{aligned}
\label{eq:G}
\end{equation}

Here, $\sigma$ denotes the softmax function, and the expected free energy balances information gain with a prior preference distribution over future outcomes or utility, encoded in $C$.

\subsection{From Free Energy to Valence}

We depart from the approach of Joffily and Coricelli~\cite{Joffily2013}, who collapse features of valence and arousal when differentiating emotional states by exclusively computing valence from variational free energy and its time derivatives. In addition, we propose a simple non-hierarchical approach to valence which, unlike the account in Hesp et al.~\cite{Hesp2021}, does not directly invoke policy selection. Instead, we derive a straightforward, psychologically interpretable description of valence in terms of the difference between the utility of observations given preferred ones and the prior expected utility. More formally, 

\begin{equation*}
\begin{aligned}
Valence\;(V) = Utility\;(U) - Expected\;Utility\;(EU),
\end{aligned}
\end{equation*}
\\
with utility of an observed observation $o_t$ at time step $t$ given by:

\begin{equation*}
\begin{aligned}
U =\log P(o_t | C)\\
\end{aligned}
\label{eq:U}
\end{equation*}
\\
and expected utility: 
\\
\begin{equation*}
\begin{aligned}
EU = \mathbb{E}_{Q(o_{t}|s_{t-1},\pi)}\big[\log P(o_t|C)]
\end{aligned}
\label{eq:EU}
\end{equation*}
\\

This formulation tracks the positive and negative experiences humans have when they do and do not encounter what they prefer to observe. The consequence is that positive valence is associated with a ``better than expected'' outcome. Such a specification of positive valence corresponds to the role of dopamine signaling in the brain, which is known to encode reward prediction errors~\cite{Schultz2016}.

\subsection{From Free Energy to Arousal}

Empirical evidence links arousal to uncertainty vis-a-vis ``increases in
amygdala activity~\cite{WilsonMendenhall2013} which can be
considered a learning signal~\cite{Li2014}'' (cited in Feldman-Barrett~\cite{Barrett2016}). Allostasis refers to an internal state of an organism which optimally predicts incoming messages in an effort to maintain homeostasis in a dynamic environment. To achieve a homeostatic state, an active agent needs to move towards an allostatic one, which amounts to updating model parameters to reduce the uncertainty of its posterior beliefs~\cite{Corcoran2020}. The empirical relationship between arousal, the amygdala, uncertainty, and the imperative to update model parameters to achieve allostasis motivates casting the entropy $H$ of a posterior distribution as an index of arousal in a dimensional model of emotion. More formally, arousal (A) becomes:
\\
\begin{equation}
\begin{aligned}
A = \mathbb{E}_{Q(s| o)}\big[-\log Q(s | o) \big] & = H[Q(s|o)]\\ 
\end{aligned}
\label{eq:F}
\end{equation}

Our interpretation of $H$ as arousal implies that posterior uncertainty is not valenced. This is in line with Feldman-Barrett's claim that arousal, associated with activity of the amygdala, is not inherently 'emotional' but is rather a signal of uncertainty~\cite{Barrett2016}. This interpretation is further motivated by the fact that there is no simple equivalence between `emotional' and `valenced' in Circumplex models, and in that model paradigmatic emotions involve both valence and arousal.

\subsection{Transformation to an Active Inference Emotional State Space}

To ground valence and arousal in a Circumplex Model, we transform the Cartesian coordinates (V, A) into the following polar coordinates. 

\begin{equation*}
\begin{aligned}
r = radius = \sqrt{V^2+A^2} & \\\theta = angle = tan^{-1}{(\frac{A}{V})}
\end{aligned}
\end{equation*}

After this transformation, we can represent agents' emotional states in a circle, spacing emotions by both degree and distance from origin, i.e., distance from neutral. The horizontal axis represents the valence dimension, ranging from sad (negative valence) on the left to happy (positive valence) on the right, while the vertical axis represents degree of arousal, ranging from high arousal/uncertainty (top) to low arousal/certainty (bottom). The distance from origin represents the ``intensity'' of the emotion (degree of affective response). The co-occurrence of valence and arousal results in the varying degrees on the circle with different emotion labels ~\cite{Posner2005}. The cited emotion labels are standardly associated with these orientations on Circumplex models.

\begin{figure}[t!]
\centering
\includegraphics[width=.45\textwidth]{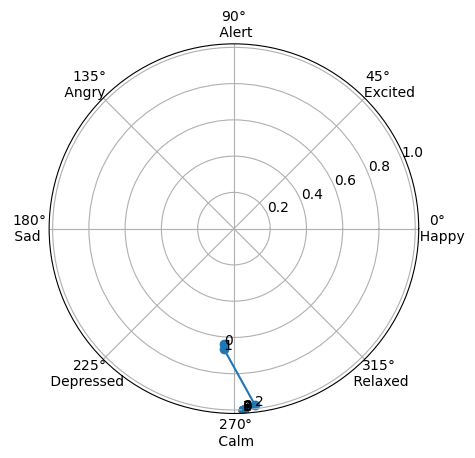}
\caption{\textbf{Free Energy Transformed Circumplex Model.} Distance to origin is emotional ``intensity'' and degree on the circle maps to different emotional states. In this case, we simulated an agent that stayed in a ``calm'' state throughout the entire trajectory.} \label{fig:circumplex}
\end{figure}

Figure 1 shows an example plot of the Circumplex Model in question. The blue trajectory describes the search sequence of a simulated agent who wants to find an arbitrary object and has a precise prior on where to find it. The resulting behavior is an agent that finds the shortest path to the expected object location and promptly locates the object at time step 2. The associated state is ``calm'' for the whole trajectory. In the next section, we will describe the simulated model in more detail, and we will will illustrate more diverse scenarios with differing resultant emotional states.

\section{Search Agent Simulation}

Imagine yourself having lost your wallet. Likely, you will immediately start searching for it. Depending on your recent memory and/or habitual behavior, you might or might not have a good idea on where to find it. Going through this process, and depending on its outcome, you will experience distinct emotional states. These are exactly the kinds of scenarios that we aim to simulate in the following experiments.

\subsection{Generative Model} 

\begin{figure}[t!]
    \centering
    \begin{subfigure}[T]{0.35\textwidth}
        \centering
        \includegraphics[width=\linewidth]{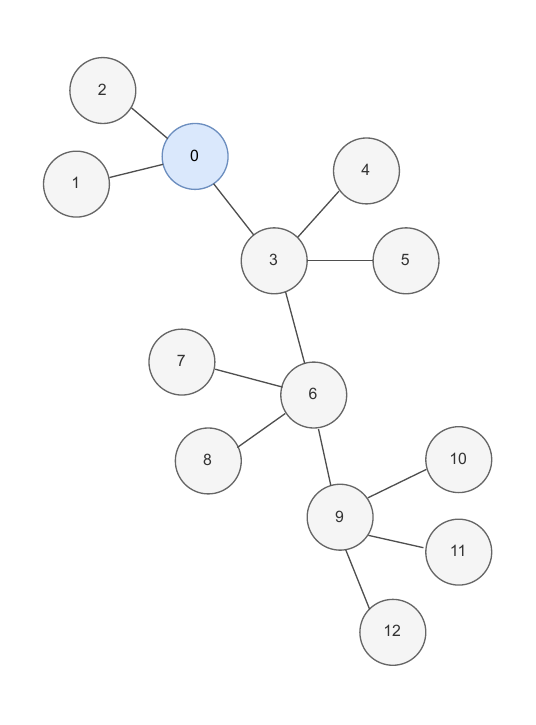}
        \caption{
        }
        \label{fig:problem-description}
    \end{subfigure}
    \begin{subfigure}[T]{0.55\textwidth}
        \centering
        \includegraphics[width=\linewidth]{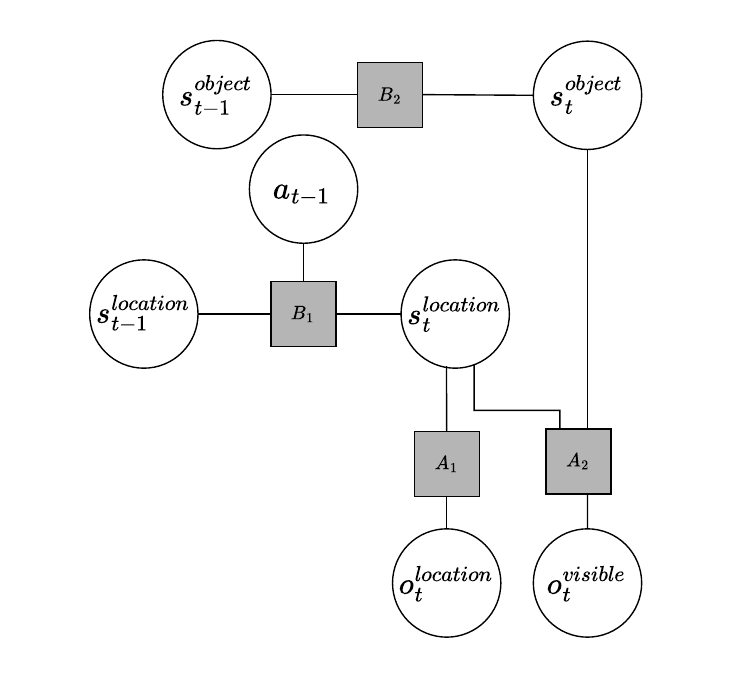}
        \caption{
        }
        \label{fig:model-detail}
    \end{subfigure}
    \caption{\textbf{Illustration of the graph environment and the agent's factor graph}. (a) Agents are located on a connected graph of locations and need to find an object that might be present at one of the locations. (b) A factor graph represents the agent's generative model. Two latent state factors that model the agent's location and the object's location, respectively, give rise to two sensory modalities through a likelihood factor: the agent's location ($A_1$) and whether the object is visible ($A_2$). The agent's location can change conditioned on move actions ($B_1$), whereas the object is kept static in our experiments ($B_2 = I$).}
    \label{fig:factor-graph}
\end{figure}

We equip our agent with the generative model represented by the factor graph in Figure~\ref{fig:factor-graph}. The agent has two state factors, one for tracking its own location, and one for maintaining a belief on the object's location. We model the environment as a connected graph of locations the agent can visit (Fig.~\ref{fig:problem-description}). At each time step, the agent can select a location to visit, but it will only transition to the next location if this is connected to its current location. These transition dynamics are encoded in $B_1$. The agent assumes the object is static and therefore the object's transition tensor is the identity matrix, i.e. $B_2=I$. The agent has two observation modalities. It can sense its current location $A_1=I$, and whether the object is visible here or not with a probability $p$, i.e. $A_2^{oij} = p$ if $i=j$ and $o=visible$. In our experiments $p=0.95$, so the agent has high confidence in seeing the object when at the correct location.

\begin{table}[b!]
    \centering
    \begin{tabular}{ c | c | c }
    \textbf{\;Scenario\;} & \textbf{\;Object Presence\;} & \textbf{\;Location Prior\;} \\
    \hline
    1 & Present     & Uniform  \\
    2 & Present     & Correct  \\
    3 & Present     & Incorrect \\
    4 & Absent      & Maybe Here \\
    5 & Absent      & Definitely Here \\
    \end{tabular}
    \vspace{0.3cm}\
    \caption{
        \textbf{Overview of search agent scenarios given levels of object presence and agent location priors.} 
        Scenario 1: Agent has uniform prior beliefs over object location. Scenario 2: Agent has a correct prior belief over object location. Scenario 3: Agent has an incorrect prior belief on object location. Scenario 4: Agent has a state dim "object not here" and object is not present. Scenario 5: Agent has a state "object must be somewhere" and object is not present. 
    }
    \label{tab:scenarios}
\end{table}

\begin{figure}[t!]
    \centering
    \includegraphics[width=.6\linewidth]{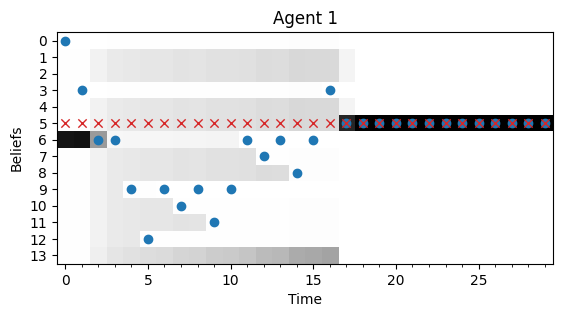}
    \caption{\textbf{Simulation of Scenario 3}. Grayscale shows the agent's belief about the object's location, whereas red x's plot the ground truth object location. The agent's own location is marked with a blue dot. In this case, the agent first has incorrect precise prior beliefs on the object location, then they do not see the object there, and they start searching other locations until it is found. }
    \label{fig:beliefs}
\end{figure}

\subsection{Design}

The agent always starts at location 0 in the graph and has a preference for the outcome ``object visible''. We simulate 5 different scenarios, varying whether the object is actually present in the environment, as well as the agent's priors. We set the agent's prior beliefs on the initial state about the object's location, which can either be uniform (i.e. no idea where the object is) or precise (i.e. remembering where the object is). In the case of a precise prior, this can be correct (i.e. the object is actually there), or not (the object is actually somewhere else, or absent). We can also equip the agent's generative model with an additional `object location` state dimension, which represents the object being absent by mapping to an invisible outcome at all locations. This can be interpreted as providing a prior belief that the object might not be present (versus that the object must be present somewhere). Table~\ref{tab:scenarios} summarizes the combinations considered per scenario.

Figure~\ref{fig:beliefs} shows the agent's actions and beliefs over time for a given scenario (Scenario 3). The belief about the object's presence at each location is plotted in grayscale, where black represents a belief of 1, indicating high certainty that the object is there, and white represents a belief of 0, correspondingly indicating high certainty that the object is not there.


\section{Results}

For each of the 5 simulation scenarios, we derived the valence and arousal values from the agent's free energy, and logged the associated emotions. In general, we found that agents were alert while searching for the object, while being happy when finding the object. Irrespective of whether the object was found, the agents in all scenarios finished in a low arousal and neutral valence state, relative to their initial emotional state. This may demonstrate their capacity for emotional regulation, or at least an inherent resistance to falling into an inescapable cycle of negative emotions.  Table 2 summarizes the trajectory of the agent in each scenario through emotional state space. We will now more completely detail the exact trajectories using the Circumplex Model as reference. 

\begin{table}[b!]
    \centering
    \begin{tabular}{ c | c }
    \textbf{Scenario\;} & \textbf{Description} \\
    \hline
    1 & Alert => Calm  \\
    2 & Calm => Calmer \\
    3 & Calm => Angry => Decreasingly alert => Relaxed => Calm\\
    4 & Alert => Calm \\
    5 & Alert => Anger <=> Depression
    \end{tabular}
    \vspace{0.3cm}\
    \caption{
        \textbf{Narrative of agent trajectory through emotional state space per scenario.} 
    }
    \label{tab:results}
\end{table}

\begin{figure}[t!]
    \centering
    \begin{subfigure}[T]{0.45\textwidth}
        \centering
        \includegraphics[width=\linewidth]{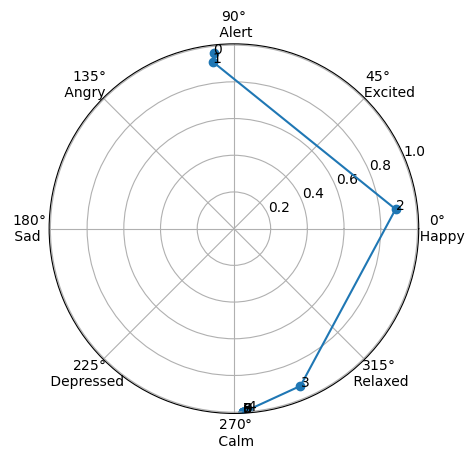}
        \caption{
        Scenario 1
        }
        \label{fig:scenario1}
    \end{subfigure}
    \begin{subfigure}[T]{0.45\textwidth}
        \centering
        \includegraphics[width=\linewidth]{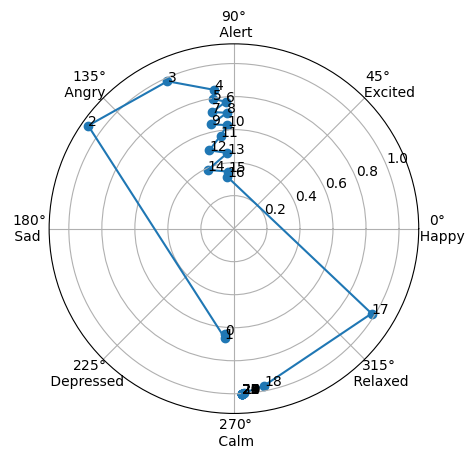}
        \caption{
        Scenario 3
        }
        \label{fig:scenario3}
    \end{subfigure}
    \caption{\textbf{Impact of Priors on Emotional State.} In Scenario 1 (left), the agent begins alert, with uniform priors. In Scenario 3 (right), the agent begins somewhere between a calm and neutral state, but they immediately become angry upon not finding the object at the location given by their prior.}
    \label{fig:results1}
\end{figure}

\subsection{Object Location Priors and Emotional State}

In the cases where the object is present and the agent ultimately locates it (Scenarios 1-3), we find that the agent begins from a calm state only when it has precise object location priors, irrespective of whether those priors ultimately prove correct. In other words, precise priors are beliefs held with high confidence, resulting in a state of ease with low arousal and neutral valence in the absence of countervailing evidence. Moreover, when the precise priors are correct (Scenario 2, shown in Figure~\ref{fig:circumplex}), the agent is calm during the entire trajectory, as it predicts that it will find the object soon from the start.

Highly valenced and aroused states are triggered only subsequent to prior assumptions being violated, which can happen either when the priors are uniform or incorrect (Fig.~\ref{fig:results1}). For instance, in Scenario 3, the violation of precise priors made the agent angry, a highly negatively valenced and positively aroused state. Subsequent to experiencing anger, when the object is located, the agent experiences a nearly 180-degree mood change into a relaxed state characterized by highly positive valence and low arousal. In Scenario 1, in which the priors are uniform (i.e. Fig.,~\ref{fig:scenario1}), the agent begins in an alert state. Interestingly, agents locating the object after beginning with uniform priors were not as relaxed when they found it as those who began with precise priors.

\subsection{Missing objects and Emotional State}


\begin{figure}[t!]
    \centering
    \begin{subfigure}[T]{0.45\textwidth}
        \centering
        \includegraphics[width=\linewidth]{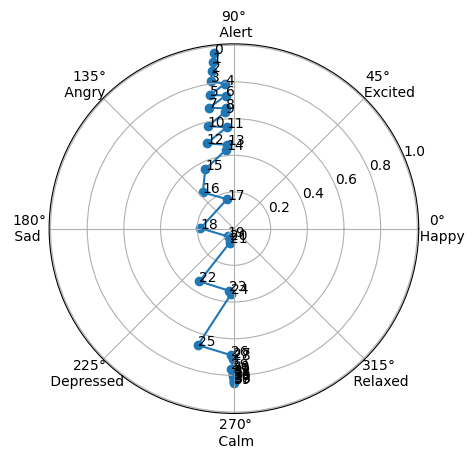}
        \caption{
        Scenario 4
        }
        \label{fig:scenario4}
    \end{subfigure}
    \begin{subfigure}[T]{0.45\textwidth}
        \centering
        \includegraphics[width=\linewidth]{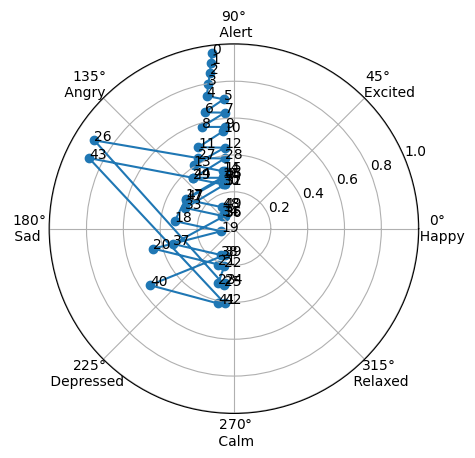}
        \caption{
        Scenario 5
        }
        \label{fig:scenario5}
    \end{subfigure}
    \caption{\textbf{Impact of object presence on Emotional State.} In Scenario 4 (left), the agent begins in the alert state, with ``maybe here'' priors. In Scenario 5 (right), the agent also begins alert, with ``definitely here'' priors. In Scenario 4 the agent ultimately accepts that there is no object to find and becomes calm, whereas in Scenario 5 the agent cycles between anger and depression when searching all locations and still not finding the object.}
    \label{fig:results2}
\end{figure}

In Scenarios 4 and 5, we simulate a search agent who will never find the object as it is not present in the environment. We conducted these simulations to see whether and how the agent conducts emotional regulation. In Scenario 4, in addition to a dimension for every location, the agent's state factor has a `not here` bin. Invoking this factor means that the agent expects not to see the object at any of the locations it can visit. In Scenario 5, the agent does not have this dimension, hence it has a structural prior that the object should always be at one of the locations. This has some interesting consequences for the emotional states the agent visits, as illustrated in the contrast between the circumplex plots in Figure~\ref{fig:results2}.

While both agents begin in the alert state, the agent in Scenario 4 (with ``maybe here'' priors) spends less time in highly negatively valenced states and ends in a state of much lower arousal upon failing to find the object. A psychological interpretation of this result is that the agent becomes successively more resigned and less interested in finding the object as time progresses, as it knows there is a possibility that the object is simply not around. In that case, not finding the object furnishes evidence. 

By contrast, the agent in Scenario 5 (with ``definitely here'' priors) spends more time in highly negatively valenced states and ends in a state of higher arousal upon failing to find the object. Interestingly, both anger and depression in this agent spike at later time steps (26 out of 43 and 20 out of 40, respectively), indicating that there is a pull towards neutral arousal that is resisted when the agent is convinced that the object must be somewhere. While this agent becomes transiently more calm as its search eliminates possibilities, providing false hope of finding the object, it ultimately cannot accept an ``object not present'' outcome, and it is in a more highly aroused state, i.e., less at peace, when the simulation terminates.

\section{Discussion}

In this paper, we derived a Circumplex Model of emotion by associating both valence and arousal with the parameters of free energy. Simulation studies provided evidence that our model enables the sound ascription of anthropomorphic emotional state trajectories to a simulated agent performing basic search tasks. These simulations further demonstrate that the inclusion of arousal as well as valence dimensions facilitates the commonsense understanding of emotion, even in simple scenarios. 

First, comparing Scenarios 1 (uniform prior) and 3 (incorrect prior), upon finding the object, both agents enter equivalently positively valenced states, but the agent in Scenario 3 finishes considerably less aroused. The emotional trajectory of the agent in Scenario 3, involving a mood swing from a highly negatively valenced state (anger) to a region of neutral valence, could not be described in terms of valence alone. 

Second, the agent in Scenario 3 became successively less aroused before finding the object. A description of emotional state in terms of valence alone would not capture the fact that after having a prior expectation violated, an agent may return to a state of alertness, which dwindles to resignation as time progresses.  

Third, Scenario 5, in which an agent's object presence prior is incorrect, is telling. It shows that on our theory, successful regulation of emotion is not inevitable for any active inference agent, but rather depends on the fundamental soundness of one's model of the world. Rigid priors that cannot change in the face of evidence may put successful allostasis, and thus optimal homeostasis and a correspondingly calm emotional state measured in terms of arousal or entropy, out of reach.

This work faces a couple of key limitations. One is that the scenarios are simple and the range of emotion evidenced is correspondingly narrow. Another is that our findings are not benchmarked against human data. Instead, we rely on face validity to make sense of the attribution of emotional states to simulated agents. A target for future experimental work is to measure how emotion impacts behavior in inference and learning, rather than merely characterizing it, following past work on emotional inference (e.g.,~\cite{Joffily2013}).

Other future research will explore the temporal dimension of emotion and emotional state attribution. Joffily and Coricelli~\cite{Joffily2013}, for example, distinguish between factive and epistemic emotions, i.e., beliefs that a state of affairs obtains, yielding an emotion such as disappointment, versus uncertain beliefs, yielding an emotion such as hope. This distinction may be naturally accounted for in terms of an agent's past or current and anticipated emotional experiences. 

Additionally, although we normalized the arousal component when transforming to polar coordinates, the raw radius of the polar coordinates may correspond to the intensity of an emotion. In line with our proposed model, some research suggests that the intensity of emotion is distinct from arousal~\cite{Raz2016}. An account of emotional intensity may have implications for other cognitive processes, such as event segmentation in an anthropomorphic account of episodic memory in simulated agents, as more intense emotional experiences are more likely to be encoded and recalled~\cite{KaYan2016}. 

Future work may explore emotions among simulated agents in social contexts. For example, finely-grained emotional inference of an agent's own emotions, as well as possession of a theory of mind, are prerequisites for empathy. Because our account leverages the arousal/valence distinction to define a granular emotional state space, it may serve as a foundation for simulating empathy in multi-agent scenarios. 

\bibliographystyle{ieeetr}
\bibliography{references}

\end{document}